\documentclass[10pt,twocolumn,letterpaper]{article}

\usepackage{ijcb}
\usepackage{times}
\usepackage[accsupp]{axessibility}
\usepackage{epsfig}
\usepackage{graphicx}
\usepackage{amsmath}
\usepackage{amssymb}
\usepackage{bm}
\usepackage{subfig}
\usepackage{caption}
\usepackage{diagbox}
\usepackage{multirow}
\usepackage{balance}

\graphicspath{{IJCB22IMAGES/}}

\usepackage[pagebackref=true,breaklinks=true,letterpaper=true,colorlinks,bookmarks=false]{hyperref}

\ijcbfinalcopy 

\ifijcbfinal\fi



\makeatletter
\newcommand*\titleheader[1]{\gdef\@titleheader{#1}}
\AtBeginDocument{%
  \let\st@red@title\@title%
  \def\@title{%
    \vskip-6.5em\bgroup\footnotesize\small\centering\@titleheader\par\egroup
    \vskip6.5em\st@red@title}
}
\makeatother

\title{Facial De-morphing: Extracting Component Faces from a Single Morph}

\titleheader{\textcolor{red}{S. Banerjee, P. Jaiswal and A. Ross, ``Facial De-morphing: Extracting Component Faces from a Single Morph," International Joint Conference in Biometrics (IJCB), Abu Dhabi UAE, 2022.}}

\begin{document}


\author{Sudipta Banerjee\\
International Institute of Information Technology Hyderabad (IIIT-H)\\
{\tt\small sudipta.b@iiit.ac.in}
\and
Prateek Jaiswal\\
International Institute of Information Technology Hyderabad (IIIT-H)\\
{\tt\small prateek.jaiswal@research.iiit.ac.in}
\and
Arun Ross\\
Michigan State University\\
{\tt\small rossarun@cse.msu.edu}
}

\maketitle
\thispagestyle{empty}

\begin{abstract}
   A face morph is created by strategically combining two or more face images corresponding to multiple identities. The intention is for the morphed image to match with multiple identities. Current morph attack detection strategies can detect morphs but cannot recover the images or identities used in creating them. The task of deducing the individual face images from a morphed face image is known as \textit{de-morphing}. Existing work in de-morphing assume the availability of a reference image pertaining to one identity in order to recover the image of the accomplice - i.e., the other identity. In this work, we propose a novel de-morphing method that can recover images of both identities simultaneously from a single morphed face image without needing a reference image or prior information about the morphing process. We propose a generative adversarial network that achieves single image-based de-morphing with a surprisingly high degree of visual realism and biometric similarity with the original face images. We demonstrate the performance of our method on landmark-based morphs and generative model-based morphs with promising results. 
\end{abstract}

\section{Introduction}
\label{Intro}
Face morphing combines multiple face images belonging to different individuals to create a composite that preserves the biometric features of all participating identities~\cite{Survey1, Survey2}. Morphs can go undetected through manual inspection and are capable of circumventing automated face recognition systems. As a result, morphed face images can be used covertly in identification documents to allow multiple individuals to gain access using a single document~\cite{Germanmorphing,NIST_2022}. Face morphing is typically done on images having ``similar" characteristics, \textit{e.g.}, similar age group, gender and ethnicity, to increase the chances of a successful attack. Following the selection of suitable individuals, their face images are aligned using landmark points. This step typically involves landmark detection followed by warping. After the image alignment process, pixels are blended using a linear interpolation scheme. The interpolated image can undergo optional post-processing such as splicing the morphed region onto the image of one of the identities or image editing such as Poisson blending to enhance the perceptual quality of the morphed image. Although landmark-based morphs are more common, recently, GAN-generated morphs are also gaining traction~\cite{MORGAN, MIP}.

The input image present in an identification document can be either morphed or non-morphed. Morph attack detection (MAD) can be performed via (i) reference-free single-image technique~\cite{SMAD1, SMAD2, SMAD3, WVURecent1} or (ii) reference-based differential-image technique~\cite{DMAD_BSIF, DMAD1, DMAD2, DMAD3, MaltoniSensors, WVURecent2}. The former detects whether an image is morphed or not using features extracted from the input document image. The latter uses a reference image to identify whether the document image is morphed or not. The reference image is typically a trusted live capture of the individual presenting the identification document at the security checkpoint. 
\begin{figure}[t]
    \centering
    \includegraphics[width=0.45\textwidth]{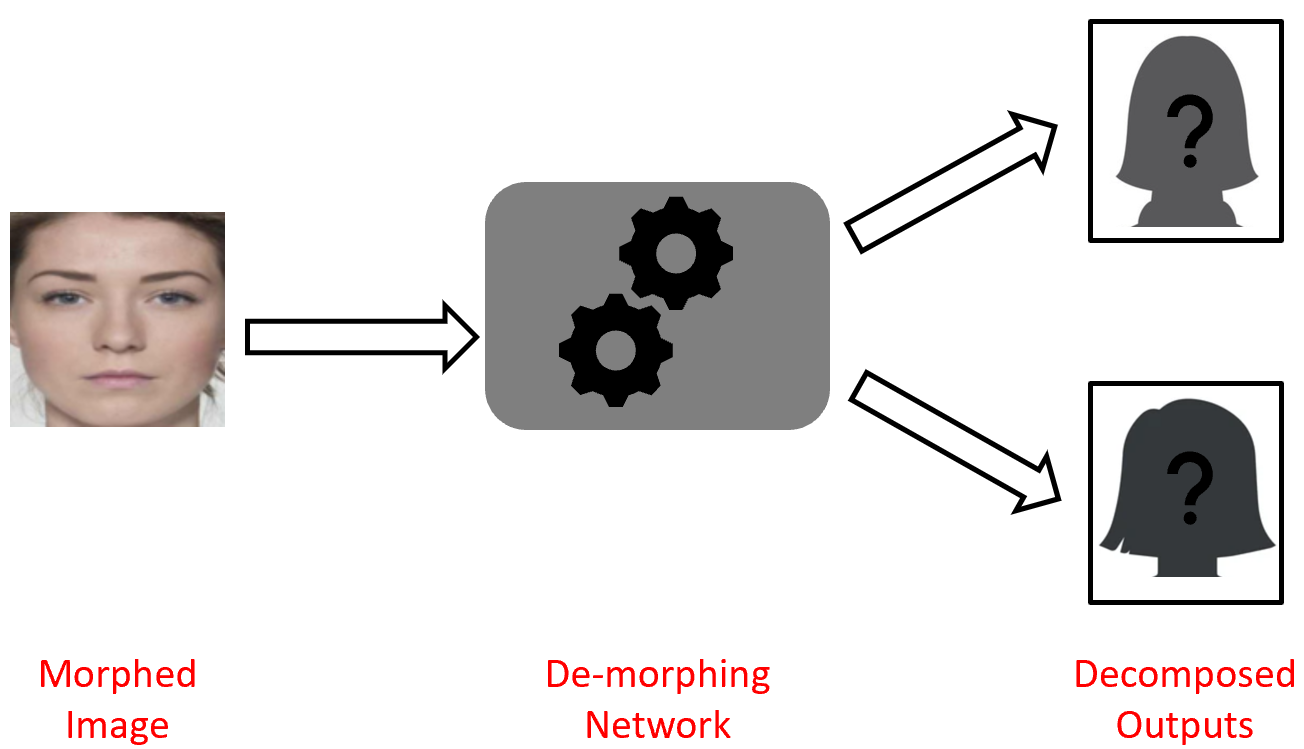}
    \caption{Objective of this work. Given a single morphed face image (example image taken from AMSL Face Morph Dataset) can we recover the two face images used in creating the morph? We propose a \textit{reference-free} de-morphing approach in contrast to existing \textit{reference-based} methods.}
    \label{fig:Obj}
\end{figure}

\begin{figure*}[h]
    \centering
    \includegraphics[width=0.8\textwidth]{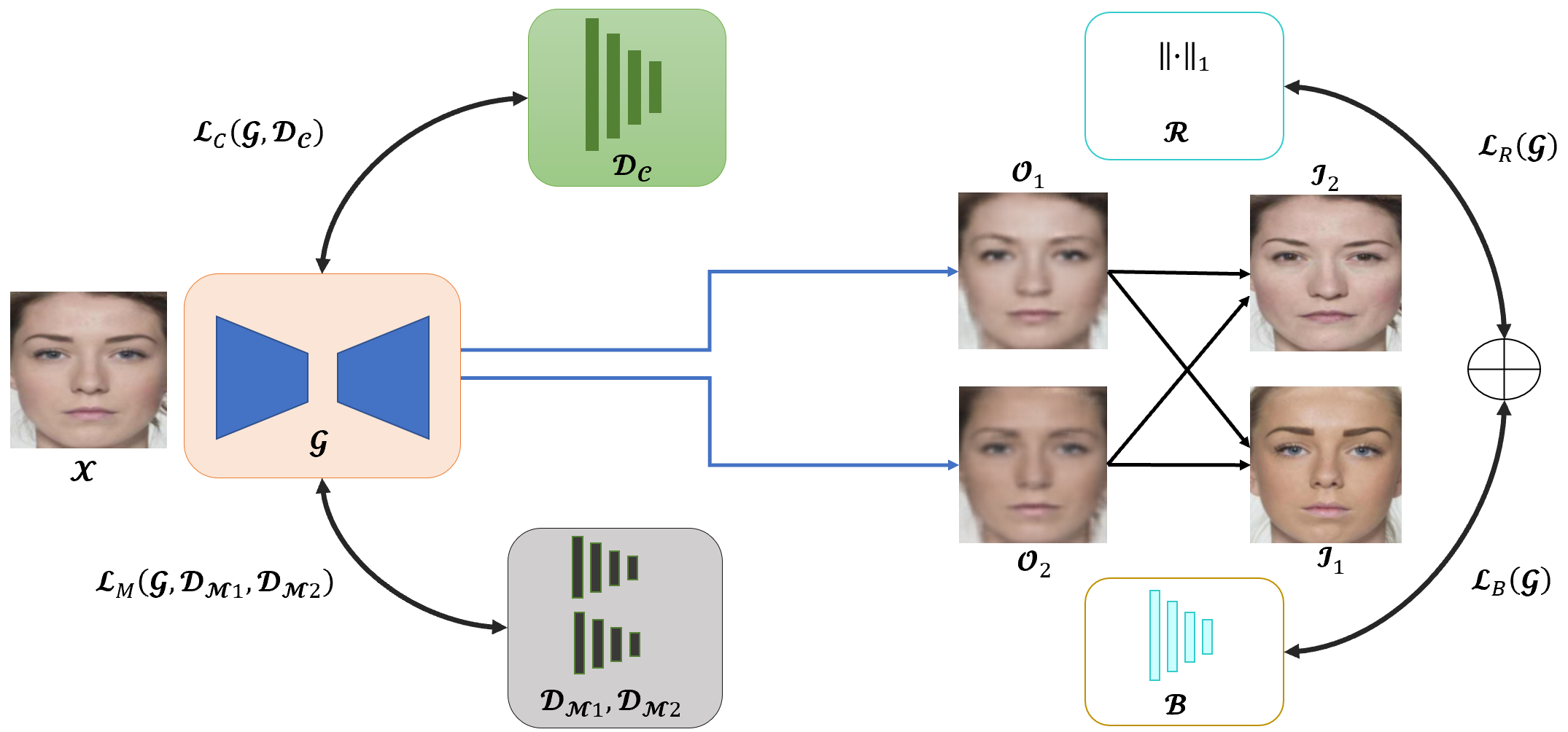}
    \caption{Overview of the proposed method. The framework decomposes a single morphed face image ($\bm{\mathcal{X}}$) using a generator ($\bm{\mathcal{G}}$), a decomposition critic ($\bm{\mathcal{D}_c}$), a pair of Markovian discriminators ($\bm{\mathcal{D}_{M1}, \mathcal{D}_{M2}}$) and a biometric representation comparator ($\bm{\mathcal{B}}$). The groundtruth input images ($\bm{\mathcal{I}_1, \mathcal{I}_2}$) supervise the adversarial training using reconstruction loss and biometric loss to generate de-morphed outputs ($\bm{\mathcal{O}_1, \mathcal{O}_2}$). The four loss terms used in this framework are: decomposition critic loss  $\bm{\mathcal{L}_C(\mathcal{G}, \mathcal{D}_C)}$, patch-based Markovian discriminator loss  $\bm{\mathcal{L}_M(\mathcal{G}, \mathcal{D}_{M_1},\mathcal{D}_{M_2})}$, cross-road reconstruction loss  $\bm{\mathcal{L}_R (\mathcal{G})}$, and cross-road biometric loss  $\bm{\mathcal{L}_B (\mathcal{G})}$.}
    \label{fig:De-Morphing}
    
\end{figure*}

Although MAD is important from a security perspective, it does not divulge information about the individuals whose images are used in creating the morph. Deciphering the identities of the subjects involved in morph creation is critical from a forensics perspective and can aid in judicial proceedings. Morph attacks can be initiated by two familiar individuals or one individual misappropriating the face image of an unsuspecting person. Therefore, the task of \textit{de-morphing}, first introduced in~\cite{DMAD3}, can recover the identities of the subjects used in creating the morph. Limited literature exists on de-morphing, and all of them require a reference image belonging to one of the subjects, say Subject1 (Subject2), to deduce the second identity, \textit{i.e.}, Subject 2 (Subject1). \textbf{In this work, our objective is to decompose or de-morph a single morphed face image into two face images, assuming that the morphed image is created using face images belonging to two individuals, without requiring any prior information about the morphing process or the identities.} To the best of our knowledge, this is the first work on face de-morphing from a single image. See Figure~\ref{fig:Obj}.

The rest of the paper is organized as follows: Section~\ref{RelWork} gives a brief overview of existing de-morphing work. Section~\ref{Prop} introduces the rationale of our work and describes the proposed method. Section~\ref{Expts} outlines the experimental protocols and describes the results. Finally, Section~\ref{Summary} concludes the paper.

\section{Related Work}
\label{RelWork}
Morphing combines two face images belonging to two distinct individuals, denoted as, $\bm{\mathcal{I}_1}$ and $\bm{\mathcal{I}_2}$, by geometrically aligning their landmarks followed by pixel-level blending to generate a morphed image, $\bm{\mathcal{X}}$, such that, 
\begin{equation}
    \bm{\mathcal{X}} = \bm{\mathcal{M}} (\bm{\mathcal{I}_1}, \bm{\mathcal{I}_2}),
\end{equation}
where, $\bm{\mathcal{M}(\cdot, \cdot)}$ denotes the morphing technique.
For the composite image to be classified as a successful morph attack, the composite should match both identities in terms of their biometric features, \textit{i.e.,} $\bm{\mathcal{B}}(\bm{\mathcal{X}},\bm{\mathcal{I}_1}) > \mathcal {T}$ and $\bm{\mathcal{B}}(\bm{\mathcal{X}},\bm{\mathcal{I}_2}) > \mathcal {T}$ (where $\bm{\mathcal{B}}$ is a biometric comparator and $\mathcal{T}$ is a user-defined threshold depending on the application scenario), and be visually reasonable in terms of perceptual quality.

 In ~\cite{DMAD3}, the authors reverse the operation of morphing to separate the second identity from the morphed image using the first identity as reference image. In order to reverse the process of morphing, the authors assumed prior knowledge about parameters of the morphing process and landmark points used in morphing. 
FD-GAN~\cite{FDGAN} utilizes a symmetric dual network and two layers of restoration losses to restore the second identity's face image while using the first identity's face image as reference. It employs a dual architecture as it first recovers the second identity's image from the morphed input using the first identity's image, and then tries to recover the first image from the morphed input using the output generated by their network. This is done to validate the effectiveness of their generative model.
In~\cite{BanerjeeRoss_IJCB2021}, the authors used a conditional GAN to perform identity disentanglement of the second subject (accomplice) from the morphed image, given the image of the first subject (anchor) as input. The conditional GAN used the pixel-wise difference image computed between the morphed input and the reference image to recover the image of the second identity via conditional entropy minimization.

All of the above methods involve the use of a reference image belonging to one of the identities, and optionally, utilize additional information about the morphing process for decoding the second identity from the morphed image. In a practical scenario, however, a reference image may not be available. 

\section{Methodology}
\label{Prop}

\subsection{Rationale}
In this paper, we propose a novel method that accepts a single morphed face image, $\bm{\mathcal{X}}$, as input, and decomposes the morphed image to produce two output images, $\bm{\mathcal{O}_1}$ and $\bm{\mathcal{O}_2}$. The two output images will belong to the two identities used in creating the morph such that $\bm{\mathcal{O}_1} \approx \bm{\mathcal{I}_1}$ and $\bm{\mathcal{O}_2} \approx \bm{\mathcal{I}_2}$ (outputs may not always be ordered). We would like to point out that our method operates on non-morphed input images also. For non-morphed input images, the method will generate two copies of the input image with slight variations, such that $\bm{\mathcal{O}_1} \approx \bm{\mathcal{O}_2} \approx \bm{\mathcal{X}}$. This decomposition task can be conceived as the very well-known blind source separation task that can be addressed using independent component analysis. However, ICA-based separation requires \textit{atleast} two observations of the input mixed signal to work effectively. In our case, we only have a single morphed image. Although augmentation techniques can be applied to generate multiple morphed images from the single input, they do not satisfy the requirement of independent observations.
Therefore, we propose a method that can take a single morphed image as input and can recover the two identities used in creating the morph. 

\subsection{Proposed Method}
In~\cite{DAD}, the authors attempt to separate the layers of a superimposed image using an adversarial network referred to as Deep Adversarial Decomposition (DAD). This finds applications in de-raining and de-hazing. However, a superimposed natural-scene image is different from a morphed face image that preserves the biometric utility of both contributing individuals. The morphing technique, $\bm{\mathcal{M}(\cdot, \cdot)}$, involves highly non-linear and complex warping and image editing operations to remove ghosting artifacts. So, deducing original identities from a single morphed input is an ill-posed problem with additional biometric constraints. We adopt the network architecture from DAD as our backbone network.   
It follows an adversarial training regime with one generator and three discriminators. The generator, $\bm{\mathcal{G}}$, acts as the de-morpher that generates the two outputs and it aims to fool the three discriminators. The first discriminator known as the decomposition critic, $\bm{\mathcal{D}_C}$, aims to distinguish between the decomposed (de-morphed) outputs and the original images used in creating the morph. A pair of Markovian discriminators, $\bm{(\mathcal{D}_{M_1}, \mathcal{D}_{M_2})}$, focus on improving high-frequency details to improve the visual realism of the generated outputs. These ``local perception networks" perform patch-based analysis to discern whether a patch belongs to an original image or a decomposed image. 

\textbf{Loss functions:} The generator and the decomposition critic are trained using the \textit{decomposition critic loss}, $\bm{\mathcal{L}_C(\mathcal{G}, \mathcal{D}_C)}$, that can be expressed as follows:
\begin{equation} \label{eq2}
\begin{split}
\bm{\mathcal{L}_C(\mathcal{G}, \mathcal{D}_C) =  \mathbb{E}_{\mathcal{I}_i \sim p_i(\mathcal{I}_i)}{[ \log \mathcal{D}_C (\mathcal{I}_1 , \mathcal{I}_2 )]}} + \\
\bm{\mathbb{E}_{\mathcal{O}_i \sim p_i(\mathcal{O}_i)}{[ \log(1 - \mathcal{D}_C (\mathcal{O}_1 , \mathcal{O}_2 ))]}} + \\
\bm{\mathbb{E}_{\mathcal{I}_i \sim p_i(\mathcal{I}_i)}{[ \log(1 - \mathcal{D}_C (\textit{mixture}(\mathcal{I}_1 , \mathcal{I}_2 )))]}} \\ i ={{1,2}}
\end{split},
\end{equation}
where, $mixture(\cdot, \cdot)$ performs a weighted linear combination of the original images to create a synthetic set that will boost the ability of the discriminator to distinguish between de-morphed images and original images.

The generator and the local perception networks are trained using \textit{patch-based Markovian discriminator loss}, $\bm{\mathcal{L}_M(\mathcal{G}, \mathcal{D}_{M_1}, \mathcal{D}_{M_2})}$, that can be expressed as follows:

\begin{equation} \label{eq3}
\begin{split}
\bm{\mathcal{L}_M(\mathcal{G}, \mathcal{D}_{M_1}, \mathcal{D}_{M_2})} = \sum_{k= 1,2} [ \bm{ \mathbb{E}_{ (\mathcal{I}_i, y)\sim p_i(\mathcal{I}_i, y)}}\\{[ \log\bm{\mathcal{D}_{M_k}(\mathcal{I}_i|y )]}} + 
\bm{\mathbb{E}_{(\mathcal{O}_i, y) \sim p_i(\mathcal{O}_i, y)}}\\{[ \log(1 - \bm{\mathcal{D}_{M_k} (\mathcal{O}_i | y))]}} ], i ={{1,2}}
\end{split},
\end{equation}

In the above equation, the conditional discriminator tries to discern whether a patch from the output of the generator, $\bm{y}$, is from the de-morphed image ($\bm{\mathcal{O}_i}$) or from the original image ($\bm{\mathcal{I}_i}$). Consider the original image as the clean image and the de-morphed image as the fake image.
In order to train the generator such that the outputs $(\bm{\mathcal{O}_1}, \bm{\mathcal{O}_2})$ closely resemble the groundtruth original images $(\bm{\mathcal{I}_1}, \bm{\mathcal{I}_2})$, in terms of perceptual quality, a  \textit{cross-road reconstruction loss}, $\bm{\mathcal{L}_R (\mathcal{G})}$ is employed. The reason for employing a ``cross-road" loss term is to account for the lack of ordering between the outputs. We cannot guarantee that $\bm{\mathcal{O}_1}$ will always correspond to $\bm{\mathcal{I}_1}$, and $\bm{\mathcal{O}_2}$ will always correspond to $\bm{\mathcal{I}_2}$. Therefore, the objective is to take all possible combinations between outputs and original inputs by interchanging the order of the outputs and then aggregating the responses. The minimum of the responses is then considered as the final loss function value. The cross-road reconstruction loss is expressed as follows:

\begin{equation} \label{eq4}
\begin{split}
\bm{\mathcal{L}_R (\mathcal{G}) =  \mathbb{E}_{\mathcal{I}_i \sim p_i(\mathcal{I}_i)}[(\mathcal{I}_1, \mathcal{I}_2), (\mathcal{O}_1,\mathcal{O}_2)]} = \\ \min [ \| \bm{\mathcal{I}_1 - \mathcal{O}_1 \|_1}  + \| \bm{\mathcal{I}_2 - \mathcal{O}_2 \|_1} , \\ \| \bm{\mathcal{I}_1 - \mathcal{O}_2 \|_1} + \| \bm{\mathcal{I}_2 - \mathcal{O}_1 \|_1}], 
\end{split}
\end{equation}

where, $\| \cdot \|_1$ indicates $l_1-norm$ and $i={1,2}$.

Although these loss functions are suitable for decomposing superimposed natural scene images, for example, removal of layer of haze from a photograph, our task is more complex. There are two fundamental differences: (1) we are dealing with face images that have structural details and possess biometric recognition utility unlike natural scene images, and (2) we are dealing with the process of morphing that subjects the images to geometrical warping, pixel intensity blending, and operations such as splicing, and removing ghosting artifacts using image editing tools. Such complex processing cannot be approximated using a single non-linear function like a gamma transformation. 

Therefore, we integrated a face matching network into our method to handle the biometric aspect of de-morphing and ensure the outputs are related to the original subjects. The face matching network accepts a pair of inputs: each of the original images used in creating the morph and each of the outputs produced by the generator. The match scores produced by the face matching network will induce a penalty for the generator during the training process if the de-morphed outputs are not comparable to the original images in terms of biometric utility. Due to the unordered nature of the outputs, we need to evaluate all possible comparisons between the original images and the generator outputs by exchanging their order. So, we introduce a new loss term known as the \textit{cross-road biometric loss}. 

The purpose of the cross-road biometric loss is to focus on the biometric utility of the output images, while the cross-road reconstruction loss focuses on the perceptual quality of the generated outputs.
The cross-road biometric loss used to train the generator is expressed as follows:

\begin{equation} \label{eq5}
\begin{split}
\bm{\mathcal{L}_B(\mathcal{G}) =  \mathbb{E}_{\mathcal{I}_i \sim p_i(\mathcal{I}_i)}[(\mathcal{I}_1, \mathcal{I}_2), (\mathcal{O}_1,\mathcal{O}_2)]} =  \min [ \bm{\mathcal{B} ( \mathcal{I}_1 ,}\\ \bm{\mathcal{O}_1 )} +  \bm{\mathcal{B} ( \mathcal{I}_2 , \mathcal{O}_2 )} , \bm{\mathcal{B (I}_1 , \mathcal{O}_2 )}  +  \bm{\mathcal{B (I}_2 , \mathcal{O}_1 )}], 
\end{split}
\end{equation}

where, $\bm{\mathcal{B}(\cdot, \cdot)}$ denotes a biometric (face) matching network that compares a pair of inputs and yields either a similarity score or a distance score, and $i=1,2$. The match score from the biometric comparator is used to determine whether the inputs belong to the same subject (genuine) or different subjects (impostor). 

The final objective function combines the four losses: decompositon critic loss (Eqn. \ref{eq2}), patch-based Markovian disciminator loss (Eqn. \ref{eq3}), cross-road reconstruction loss (Eqn. \ref{eq4}) and cross-road biometric loss (Eqn. \ref{eq5}) as follows:

\begin{equation} \label{eq6}
\begin{split}
\bm{\mathcal{L} (\mathcal{G}, \mathcal{D}_C, \mathcal{D}_{M_1}, \mathcal{D}_{M_2})} = \bm{\mathcal{L}_R ( \mathcal{G})}  + \beta_C \bm{\mathcal{L}_C (\mathcal{G}, \mathcal{D}_C)} \\ +  \beta_B \bm{\mathcal{L}_B (\mathcal{G})} + \beta_M \bm{\mathcal{L}_M (\mathcal{G}, \mathcal{D}_{M_1}, \mathcal{D}_{M_2} )}. 
\end{split}
\end{equation}

In Eqn. \ref{eq6}, $\beta_C, \beta_B, \beta_M$ are regularization parameters that will be discussed in implementation details. See Figure~\ref{fig:De-Morphing}.

\section{Experiments and Results}
\label{Expts}
\subsection{Dataset}
\label{Data}


 We used three face morph datasets in this work. (i) \textbf{AMSL face morph dataset}~\cite{AMSL1, AMSL2}: It contains images from 102 subjects captured with neutral as well as smiling expressions. There are 2,175 morphed images corresponding to 92 subjects created using a landmark-based approach. Some of the non-morphed images have not been used to create morphs. The train and test splits are done to follow subject disjoint protocol following an approximate 70:30 split in terms of subjects between train and test sets. (ii) \textbf{E-MorGAN dataset}~\cite{CIEMORGAN}: It contains 498 non-morphed images. Two morphs are created using a generative network and a cascaded image enhancement network to improve visual realism from the two subjects most similar to each of the non-morphed images resulting in 1,000 morphed images, that were further split into train and test sets. We used the entire train split of morphed images provided by the original authors and a subset of the test split to maintain balance between morphed and non-morphed images in the test set. (iii) \textbf{ReGenMorph dataset}~\cite{ReGen}: It contains 2,500 morphed images and 1,270 bonafide images. Face images from the FRGCv2.0 dataset are used to first create the morphs using a landmark-based face morphing approach to reduce the blending artifacts and then fed to a StyleGAN network. The generator processes the morphs initially created using a landmark-based approach in the latent space to produce realistic morph attacks.
 The final train and test split for all three datasets is presented in Table~\ref{Tab1:AMSLData}. In our work, we used only morphed images for training but both morphed and non-morphed images for testing.
 

\begin{table}[h]
\centering
\caption{Dataset specifications.}
\scalebox{0.85}{
\begin{tabular}{|l|l|l|l|l|}
\hline
\textbf{Dataset}        & \textbf{Split}        & \textbf{\begin{tabular}[c]{@{}l@{}}Morphed/\\ Non-morphed\end{tabular}} & \textbf{\begin{tabular}[c]{@{}l@{}}No. of \\ subjects\end{tabular}} & \textbf{\begin{tabular}[c]{@{}l@{}}No. of \\ images\end{tabular}} \\ \hline \hline
\multirow{3}{*}{AMSL}   & Train                 & Morphed                                                                 & 73                                                                  & 946                                                               \\ \cline{2-5}
                        & \multirow{2}{*}{Test} & Morphed                                                                 & 16                                                                  & 56                                                                \\
                        &                       & Non-morphed                                                             & 29                                                                  & 57                                                                \\ \hline
\multirow{3}{*}{E-MorGAN} & Train                 & Morphed                                                                 &     251                                                                &      499                                                            \\ \cline{2-5}
                        & \multirow{2}{*}{Test} & Morphed                                                                 &  90                                                                   &      100                                                             \\
                        &                       & Non-morphed                                                             &   50                                                                  &  100 \\ \hline       
\multirow{3}{*}{ReGenMorph} & Train                 & Morphed                                                                 &     93                                                                &      1,008                                                            \\ \cline{2-5}
                        & \multirow{2}{*}{Test} & Morphed                                                                 &  40                                                                  &      238                                                           \\
                        &                       & Non-morphed                                                             &   40                                                                 &  40 \\ \hline       
\end{tabular}}
\label{Tab1:AMSLData}
\end{table}

                    

\subsection{Implementation details}
\label{imp}
We cropped the images to detect and include only the face region using MTCNN~\cite{MTCNN}. To implement the generator, we used the U-Net architecture. The decomposition critic is a 4-layer fully convolutional network. The Markovian discriminators are a pair of 3-layer fully convolutional networks following the implementation details provided by~\cite{DAD}. We built our method by adopting the deep adversarial decomposition architecture developed by the original authors of~\cite{DAD}. We used the open-source implementation~\cite{DeepFace_Github} of the ArcFace~\cite{ArcFace} network to implement our biometric comparator with cosine distance as biometric score. Training was done using Adam optimization. Training parameters were as follows: Number of epochs=300, $\beta_C = \beta_M = 0$ for first 10 epochs, $\beta_C = \beta_M = 0.001$ for the remaining 290 epochs, $\beta_B  = 10^{12}$ and learning rate = $10^{-4}$. 

\subsection{Findings}
\label{Findings}
\begin{figure}[h]
    \centering
    \includegraphics[width=0.43\textwidth]{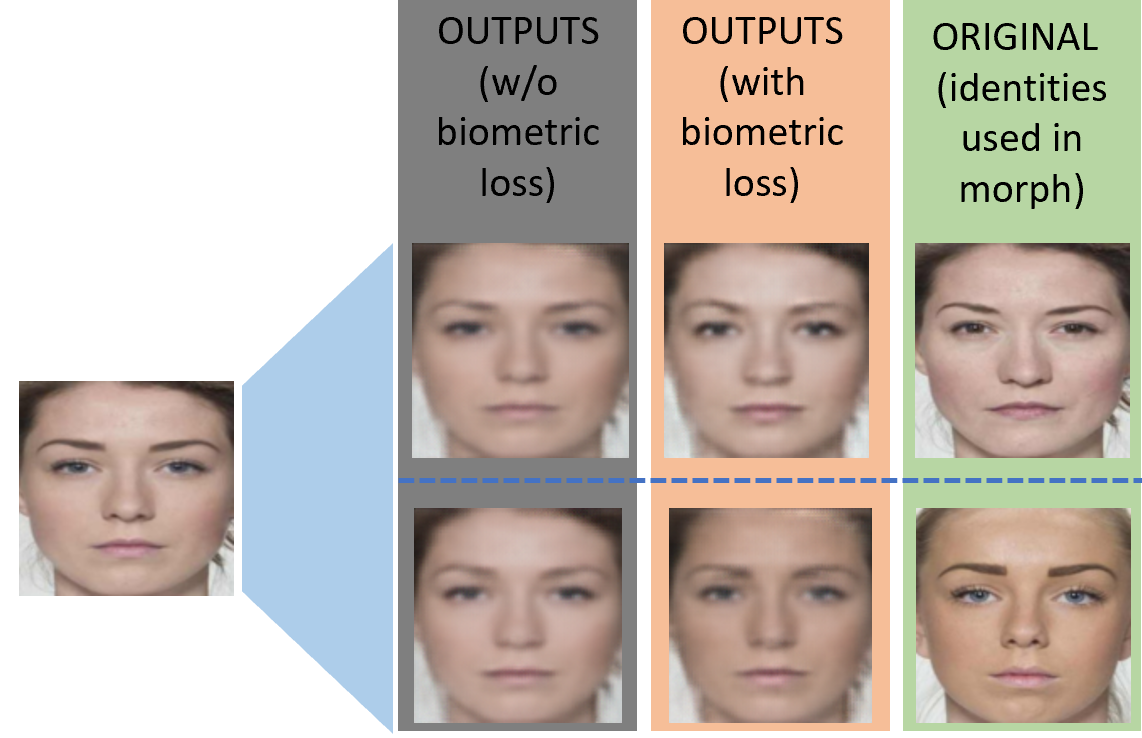}
    \caption{Qualitative visualization of the impact of biometric loss $\bm{\mathcal{L}_{B}}$ on the de-morphed outputs from the morphed input image presented on the left. The first column represents de-morphed outputs \textit{without} biometric loss, the second column represents de-morphed outputs \textit{with} biometric loss, and the third column represents the original face images used in creating the morph. Note the positive role played by the biometric loss in producing well-separable de-morphed outputs.}
    \label{fig:Biomloss}
\end{figure}

 We evaluated the proposed method on \textit{both} morphed and non-morphed images to validate that our method is generating outputs that are consistent with the input. Ideally, non-morphed input should yield identical outputs as only one identity is assumed to be present in the input image. In order to evaluate the importance of the cross-road biometric loss term in the proposed method, we presented an illustration in Figure~\ref{fig:Biomloss}. The de-morphed outputs from the single morphed image appear to be near-duplicates in the \textit{absence} of the biometric loss term (first column). The visual discriminability between the outputs is more pronounced in the presence of the biometric loss (second column). Biometric loss helps the network disentangle identity-specific features to successfully recover the original face images (third column). 

\begin{figure*}[h]
    \centering
    \includegraphics[width=0.57\textwidth]{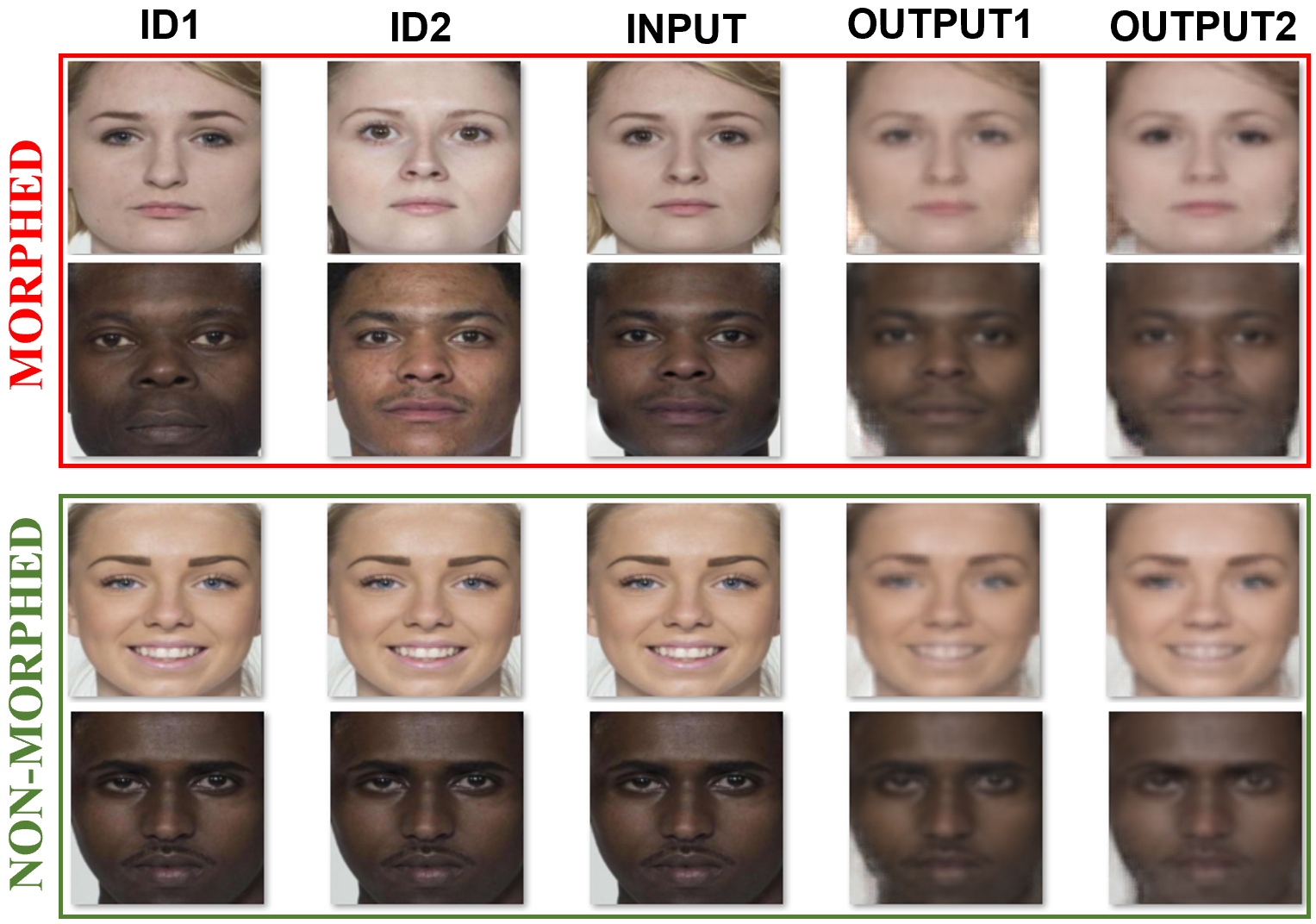}
    \caption{Examples of de-morphed images generated using the proposed method on the AMSL Face Morph dataset. Here, ``INPUT" represents \textbf{morphed} document image in the first two rows and \textbf{non-morphed} document image in the last two rows. ``ID1" and ``ID2" represent the original face images used in creating the morphed input image in the first two rows, and duplicates of the input image in the last two rows. ``OUTPUT1" and ``OUTPUT2" are the de-morphed outputs generated by our method. Note that the outputs are un-ordered and our method is not trained on non-morphed images but can produce realistic face images nonetheless.}
    \label{fig:MorphAMSL_Examples}
\end{figure*}

\begin{figure*}[h]
    \centering
    \includegraphics[width=0.57\textwidth]{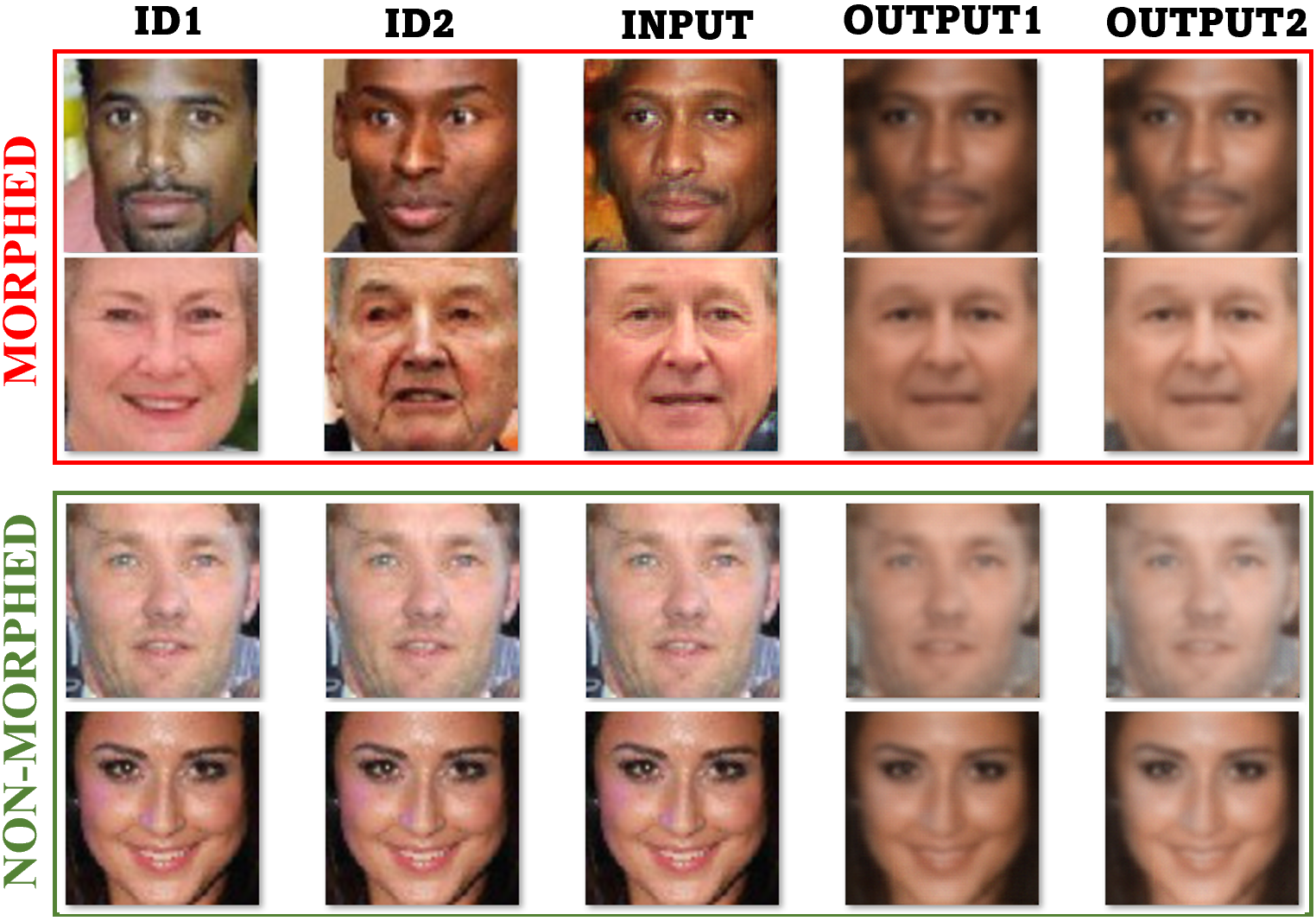}
    \caption{Examples of de-morphed images generated using the proposed method on the E-MorGAN dataset. Here, ``INPUT" represents \textbf{morphed} document image in the first two rows and \textbf{non-morphed} document image in the last two rows. ``ID1" and ``ID2" represent the original face images used in creating the morphed input image in the first two rows, and duplicates of the input image in the last two rows. ``OUTPUT1" and ``OUTPUT2" are the de-morphed outputs generated by our method.}
    \label{fig:MorphEMORGAN_Examples}
\end{figure*}

\begin{figure*}[h]
    \centering
    \includegraphics[width=0.57\textwidth]{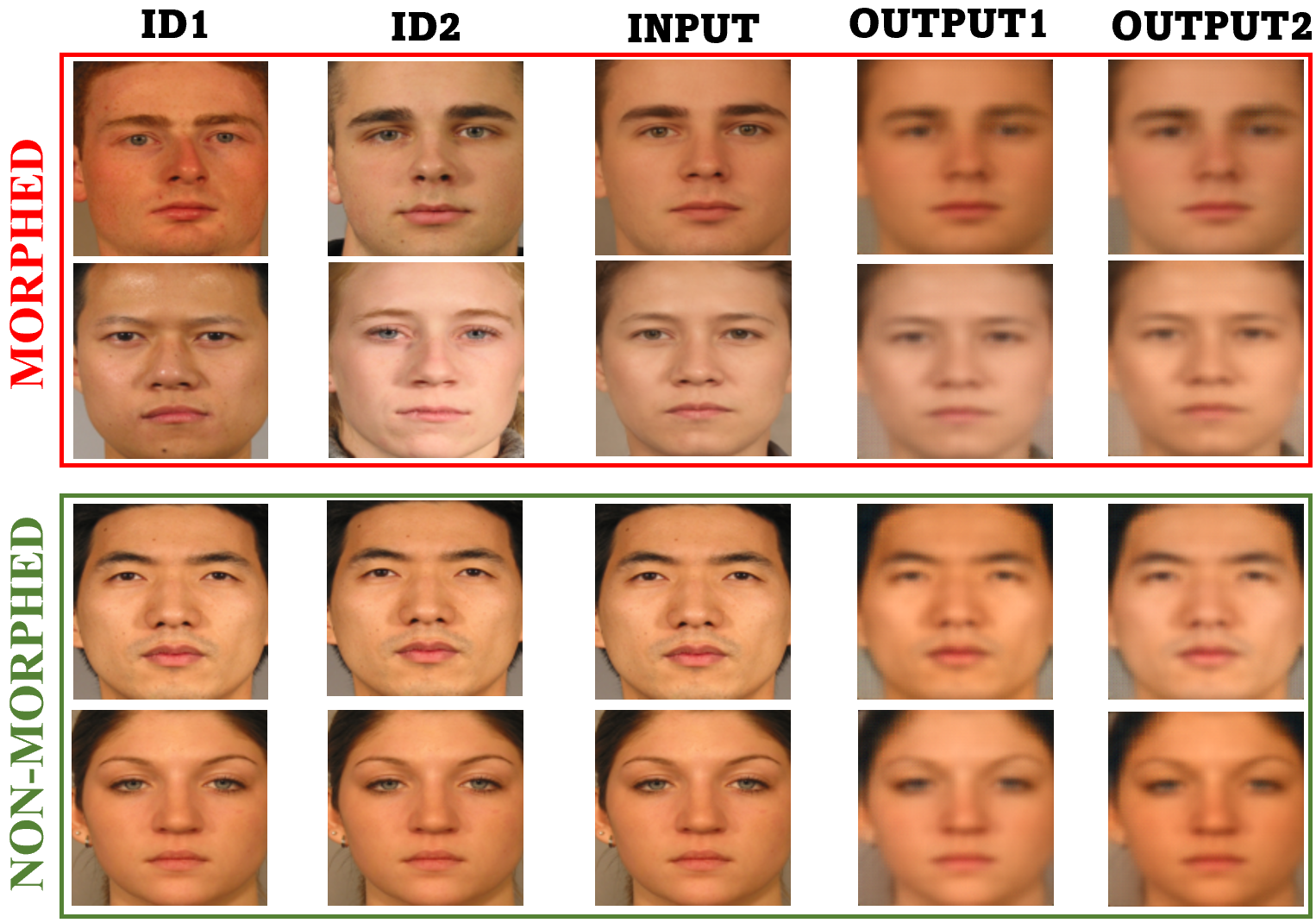}
    \caption{Examples of de-morphed images generated using the proposed method on the ReGenMorphs dataset. Here, ``INPUT" represents \textbf{morphed} document image in the first two rows and \textbf{non-morphed} document image in the last two rows. ``ID1" and ``ID2" represent the original face images used in creating the morphed input image in the first two rows, and duplicates of the input image in the last two rows. ``OUTPUT1" and ``OUTPUT2" are the de-morphed outputs generated by our method.}
    \label{fig:MorphReGen_Examples}
\end{figure*}

We performed both qualitative and quantitative evaluation of our de-morphing method. See Figure~\ref{fig:MorphAMSL_Examples} to visualize the outputs produced by the proposed method on the AMSL dataset, Figure~\ref{fig:MorphEMORGAN_Examples} to visualize the outputs on the E-MorGAN dataset and Figure~\ref{fig:MorphReGen_Examples} to visualize the outputs on the ReGenMorphs dataset. In Figures~\ref{fig:MorphAMSL_Examples}, ~\ref{fig:MorphEMORGAN_Examples} and~\ref{fig:MorphReGen_Examples}, we present two examples of \textbf{morphed} images in the first two rows from the AMSL, E-MorGAN and ReGenMorphs datasets, respectively, denoted by the third column (INPUT). The first two columns (ID1 and ID2) correspond to the original images used in creating the morph. The last two columns (OUTPUT1 and OUTPUT2) correspond to the outputs produced by the de-morphing method. Note that ID1 is the anchor image (background) that is used in creating the morph. So, the outputs mimic the features such as hairstyle of the anchor subject. However, the distinction between the outputs generated by our method is apparent in terms of variations in the skin color, and subtle details such as eyebrows and facial hair. In the last two rows of Figures~\ref{fig:MorphAMSL_Examples}, ~\ref{fig:MorphEMORGAN_Examples} and~\ref{fig:MorphReGen_Examples}, we present two examples of \textbf{non-morphed} images from the AMSL, E-MorGAN and ReGenMorphs datasets, respectively, denoted by the third column (INPUT). The first two columns (ID1 and ID2) correspond to duplicates of non-morphed image for quantitative evaluation. The last two columns (OUTPUT1 and OUTPUT2) correspond to the outputs produced by the de-morphing method. \textbf{Note that the algorithm was not trained on non-morphed images}. However, the method is able to generate two very similar looking outputs while preserving the facial expression in the input. 

\begin{figure*}[t]
    \centering
    \subfloat[Morphed]
    {
    \centering
    \includegraphics[scale=0.462]{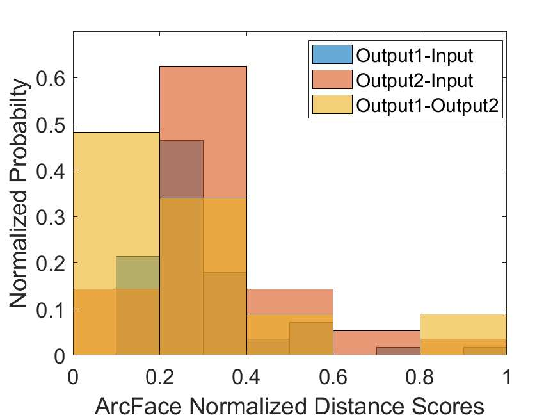}
    }
    \subfloat[Non-morphed]
    {
    \centering
    \includegraphics[scale=0.462]{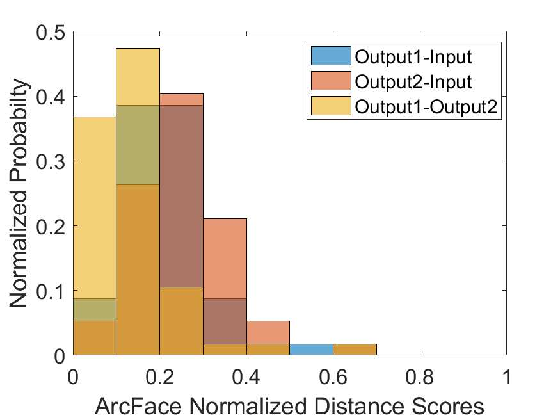}
    }
    \caption{Variations in the distance score distributions between input images (morphed and non-morphed) and outputs generated using the proposed method using ArcFace comparator on the AMSL dataset. (a) Morphed, and (b) Non-morphed.}
    \label{fig:Histograms_AMSL}
\end{figure*}

\begin{figure*}[t]
    \centering
    \subfloat[Morphed]
    {
    \centering
    \includegraphics[scale=0.462]{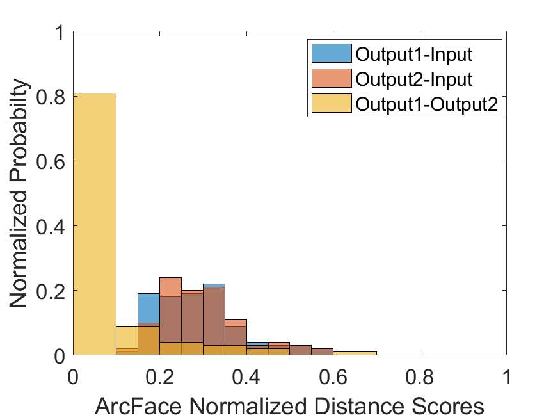}
    }
    \subfloat[Non-morphed]
    {
    \centering
    \includegraphics[scale=0.462]{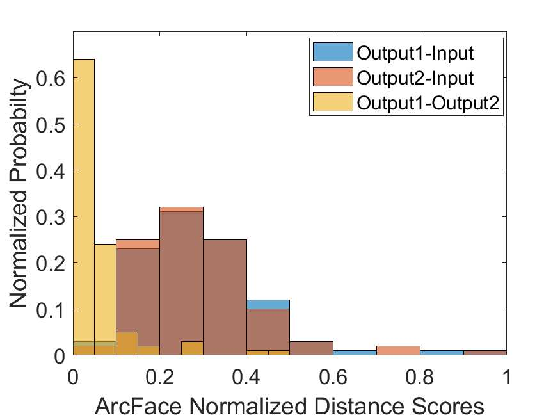}
    }
    \caption{Variations in the distance score distributions between input images (morphed and non-morphed) and outputs generated using the proposed method using ArcFace comparator on the E-MorGAN dataset. (a) Morphed, and (b) Non-morphed.}
    \label{fig:Histograms_EMORGAN}
\end{figure*}

\begin{figure*}[t]
    \centering
    \subfloat[Morphed]
    {
    \centering
    \includegraphics[scale=0.462]{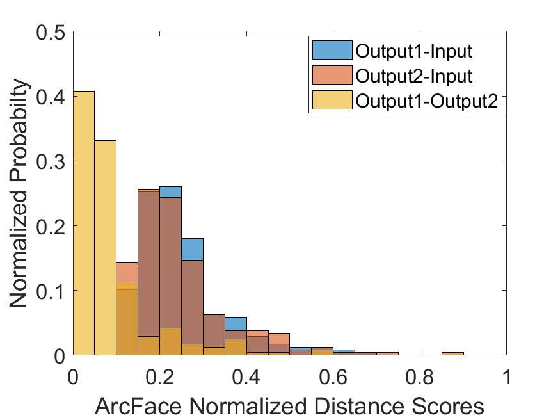}
    }
    \subfloat[Non-morphed]
    {
    \centering
    \includegraphics[scale=0.462]{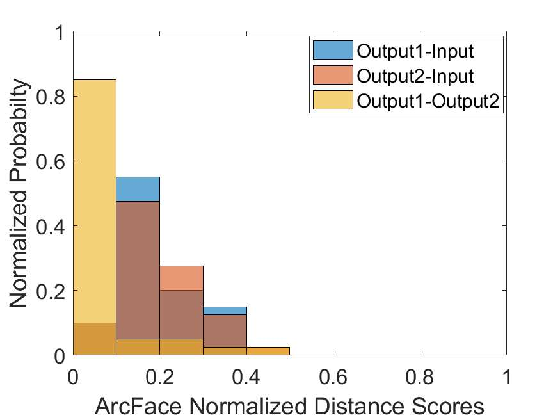}
    }
    \caption{Variations in the distance score distributions between input images (morphed and non-morphed) and outputs generated using the proposed method using ArcFace comparator on the ReGenMorphs dataset. (a) Morphed, and (b) Non-morphed.}
    \label{fig:Histograms_ReGenMorphs}
\end{figure*}

We used the ArcFace comparator and cosine distance to compute the biometric distance between the output images and the input images. We repeated this exercise for morphed and non-morphed images. The reason for doing this experiment was to ensure that the GAN was not generating arbitrary face images. Instead, the outputs produced by the GAN match in terms of biometric utility with the morphed input image. This will be true only if the outputs are related to the real subjects that have been used in creating the morph. Figures~\ref{fig:Histograms_AMSL}(a), ~\ref{fig:Histograms_EMORGAN}(a) and ~\ref{fig:Histograms_ReGenMorphs}(a) present the score distributions for \textbf{morphed} images from the AMSL. E-MorGAN and RegenMorphs datasets, respectively. Similarly, Figures~\ref{fig:Histograms_AMSL}(b), ~\ref{fig:Histograms_EMORGAN}(b) and ~\ref{fig:Histograms_ReGenMorphs}(b) present the score distributions for \textbf{non-morphed} images from the AMSL. E-MorGAN and RegenMorphs datasets, respectively. Additionally, we computed two distributions, average of the distance scores between the outputs and the input for (i) morphed images and (ii) non-morphed images, denoted as $\bigg(\frac{\bm{\mathcal{B}(Output1, Input)}+\bm{\mathcal{B}(Output2, Input)}}{2} \bigg)$. Next, we computed the \textit{d-prime} value using these two distributions. We observed it to be 4.25 for AMSL, 0.60 for E-MorGAN and 4.07 for ReGenMorphs. A high d-prime value indicates that the two distributions can be well-separated, and therefore, be potentially used for separating morphs from non-morphs.


 Finally, we compared the outputs with the groundtruth identities, \textit{i.e.}, the original identities used in creating the morph (ID1 and ID2) using a COTS face comparator that produces biometric similarity score normalized to [0,1]. As the outputs were unordered, we computed the scores with all four possible combinations: ID1-OUTPUT1, ID2-OUTPUT2, ID1-OUTPUT2 and ID2-OUTPUT1. We then adopted the following decision rule. If the sum of similarity scores between ID1-OUTPUT1 and ID2-OUTPUT2 is \textit{higher} than the sum of scores between ID1-OUTPUT2 and ID2-OUTPUT1, we considered (ID1,OUTPUT1) and (ID2,OUTPUT2) as genuine pairs. If the sum of similarity scores between ID1-OUTPUT1 and ID2-OUTPUT2 is \textit{lower} than the sum of scores between ID1-OUTPUT2 and ID2-OUTPUT1, we considered (ID2,OUTPUT1) and (ID1,OUTPUT2) as genuine pairs. Using this procedure we achieved the following results for morphed images. We report the results in terms of True Match Rate (TMR) at a False Match Rate (FMR)=10\%. (i) \textbf{AMSL}: TMR=62.5\%  for Subject 1 and a TMR=78.6\% for Subject 2, (ii) \textbf{E-MorGAN}: TMR=53.0\% for Subject 1 and a TMR=51.0\% for Subject 2, and (iii) \textbf{ReGenMorphs}: TMR=90.7\% for Subject 1 and a TMR=90.3\% for Subject 2. For non-morphed images the TMR was observed to be close to 100\% for AMSL and ReGenMorphs datasets, and TMR=60\% for E-MorGAN dataset. The results indicate that the proposed method performs reasonably well on high-quality morphs such as the AMSL and ReGenMorphs datasets compared to the E-MorGAN dataset that uses unconstrained source images for constructing the morphs. Note we used a relatively high FMR of 10\%, but we would like to emphasize the high potential of the proposed method for solving the difficult task of de-morphing from a single image. 
 


\section{Summary}
\label{Summary}
In this work, we propose a de-morphing method to recover face images belonging to two identities simultaneously from a single morphed face image. Our method does not need any prior information about the morphing process or any reference image that is typically required by existing de-morphing strategies. We used a generative model with one generator and three discriminators to recover the identities (images) used in creating the morph. We incorporated a cross-road biometric loss term to train the network to decouple the identity-specific features that are combined non-trivially in the morphed image. We presented our findings by evaluating our method on AMSL, E-MorGAN and ReGenMorphs datasets with visually compelling results and reasonable biometric verification performance with original face images. Future work will focus on developing attention guided de-morphing network to improve performance.

\section*{Acknowledgment}
The authors would like to thank Protichi Basak for computing the COTS face comparator scores.

\balance
{
\bibliographystyle{ieee}
\bibliography{Main}
}

\end{document}